\documentclass{article}
\usepackage[final]{graphicx}
\usepackage{authblk}
\usepackage{url}

\begin{document}
\title{Multi-stage Neural Networks with Single-sided Classifiers for False Positive Reduction and its Evaluation using Lung X-ray CT Images}

\author[1]{Masaharu Sakamoto}
\author[1]{Hiroki Nakano}
\author[2]{Kun Zhao}
\author[2]{Taro Sekiyama}

\affil[1]{IBM Tokyo Laboratory, Watson Health, Tokyo, 103-8510, Japan}
\affil[2]{IBM Research – Tokyo, Tokyo, 103-8510, Japan}

\maketitle

\begin{abstract}
Lung nodule classification is a class imbalanced problem because nodules are found with much lower frequency than non-nodules. In the class imbalanced problem, conventional classifiers tend to be overwhelmed by the majority class and ignore the minority class. We therefore propose cascaded convolutional neural networks to cope with the class imbalanced problem. In the proposed approach, multi-stage convolutional neural networks that perform as single-sided classifiers filter out obvious non-nodules. Successively, a convolutional neural network trained with a balanced data set calculates nodule probabilities. The proposed method achieved the sensitivity of 92.4\% and 94.5\% at 4 and 8 false positives per scan in Free Receiver Operating Characteristics (FROC) curve analysis, respectively.
\end{abstract}
\section{Introduction}
Lung cancer occupies a high percentage in the mortality rates of cancer even on a worldwide basis \cite{cancerresearch}. Early detection is one of the most promising strategies to reduce lung cancer mortality \cite{van2010comparing}. In recent years, along with performance improvements of CT equipment, increasingly large numbers of tomographic images have come to be taken (e.g., at slice intervals of 1 mm), resulting in improvements in the ability of radiologists to distinguish nodules. However, there is a limitation to interpreting a large number of images (e.g., 300 - 500 slices / scan) by relying on humans. Computer-aided diagnosis (CAD) systems show promise for the urgent task of time-efficient interpretation of CT scans. In one study \cite{van2010comparing}, six computer-aided diagnosis algorithms of lung nodules in computed tomography scans were compared. These methods extract features in lung nodule images with a signal processing technique and classify nodule candidates by using pattern matching based on statistics or a machine learning method such as the k-nearest neighbor algorithm (k-NN) and neural networks. By combining six computer-aided diagnosis algorithms, they obtained detection sensitivities of 81.6\% and 87.0\% at 4 and 8 false positives per scan in Free Receiver Operating Characteristics (FROC) curve, respectively. In recent years, spurred by the large amounts of available data and computational power, Convolutional Neural Network (CNN) has outperformed state-of-the-art techniques in several computer vision applications \cite{krizhevsky2012imagenet}. This is because CNN can be trained end-to-end in a supervised fashion while learning highly discriminative features, thus removing the need for handcrafting nodule feature descriptors. Setio, et al. \cite{setio2016pulmonary} used a CNN specifically trained for lung nodule detection. On 888 scans of a publicly available dataset (the dataset is the same as we use in this study.), their method reached high detection sensitivities of 90.1\% and 91.5\% at 4 and 8 false positives per scan in FROC curve, respectively. Dou et al. \cite{dou2016multi} proposed a method employing 3D CNNs for false positive reduction in automated pulmonary nodule detection from volumetric CT scans. 

Lung nodule classification is a class imbalanced problem, as nodules are found with much lower frequency than non-nodules. In other words, many irregular lesions that are visible in CT images are non-nodules, such as blood vessels or ribs. In the class imbalanced problem, conventional classifiers tend to be overwhelmed by the majority class and ignore the minority class. Several approaches have proposed to deal with the problems in the rare medical diagnosis \cite{rahman2013addressing}, detection of oil spills in satellite radar images \cite{kubat1998machine} and the detection of fraudulent calls \cite{fawcett1997adaptive}. Japkowicz \cite{japkowicz2000learning} showed that oversampling the minority class and subsampling the majority class are both very effective methods of coping with the problem. Chawla et al. \cite{chawla2002smote} proposed SMOTE (Synthetic Minority Over-sampling Technique) algorithm that is artificially creating minor class and randomly sub-sample majority class. Kubat and Matwin \cite{Kubat97addressingthe} proposed a one-sided selection method that keeps all minor class samples and subsamples the majority class samples. Sun et al. \cite{sun2009classification} reviewed comprehensively the class imbalanced problems. 

\begin{figure}[htb]
\centering
\includegraphics[width=\textwidth]{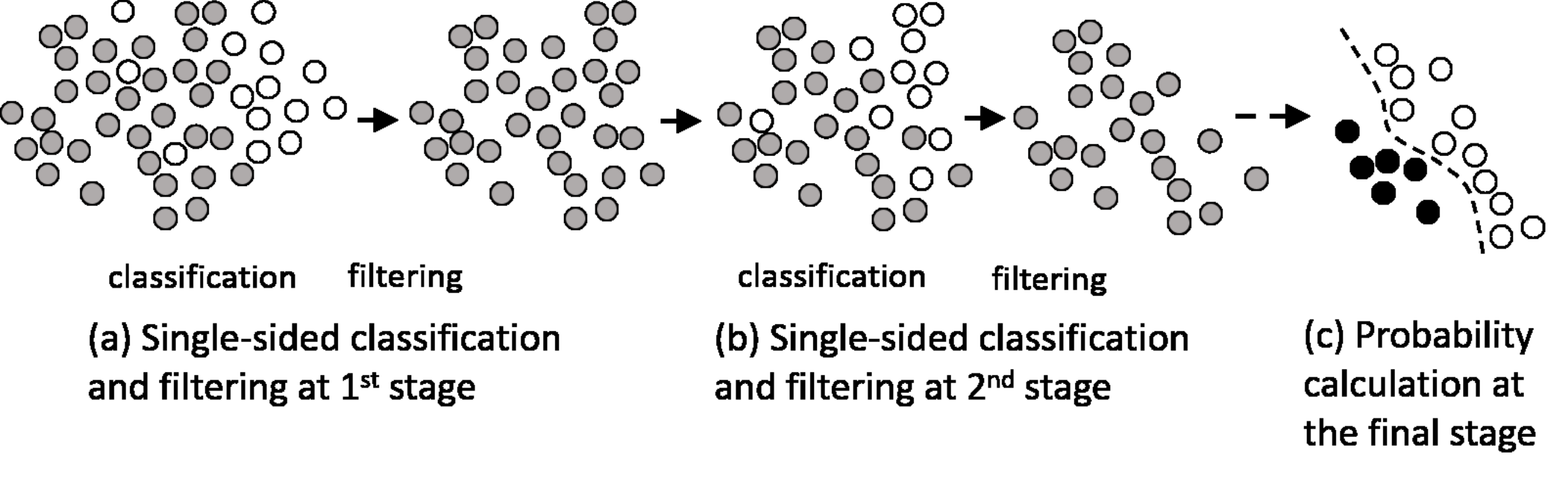}
\caption{Multi-stage processing with single-sided classifiers. It classifies the samples as suspicious nodule candidates (gray circles) and obvious non-nodules (while circles). The obvious non-nodules are filtered out st each stage. At the final stage, nodule probabilities are calculated.}
\label{fig:single-sided}
\end{figure}

As one method to cope with the class imbalanced problem in lung nodule classification, we propose a filtering method to take off majority class samples from test dataset. Our method is completely different from previous methods. It positively utilizes deterioration of classification performance caused by class imbalance learning. We call such classifiers as single-sided classifiers because it filters out majority class samples only. The single-sided classifier consists of a CNN that outputs nodule probability and a filter that removes the majority class samples by using a threshold in nodules probability. It has two kinds of outputs: the obvious non-nodules and suspicious nodule candidates. To implement such classifiers, the CNNs are trained with an inversed imbalanced dataset consisting of many nodule images and a few non-nodule images. By “inverse” we mean that the ratio of the number of nodules and non-nodules is reversed against the original dataset. As the results, the single-sided classifiers work well for nodule samples, but not work well for non-nodule samples. By using a threshold operation in nodule probability, the non-nodule samples are classified into obvious non-nodules and suspicious nodule candidates. In addition, the single-sided classifiers are concatenated in cascade arrangement. The obvious non-nodules are dismissed and assigned zero probability, the suspicious nodule candidates are passed to the down-stream classifiers. This filtering mechanism contribute to false positive reduction. Figure \ref{fig:single-sided} shows an illustration of our method. The obvious non-nodules (white circles) are filtered out at each stage, finally suspicious nodule candidates (gray circles) remain.
The aim of our method is not to balance the number of samples in majority and minority class, we just want to filter out what is a nodule without any doubt. In the final stage, the CNN trained by a balanced dataset extracted from the suspicious nodule candidates calculate nodule probabilities. By “balanced” we mean that the number of nodules is almost equal to the number of non-nodules. We rely on the CNNs which have excellent classification ability to calculate nodule probabilities of suspicious nodule candidates. As the result, our method can achieve low false positives. It helps decreasing the burden of image interpretation on radiologists. 

\section{Multi-stage Neural Networks with Single-sided Classifiers}
\begin{figure}[ht]
\centering
\includegraphics[width=\textwidth]{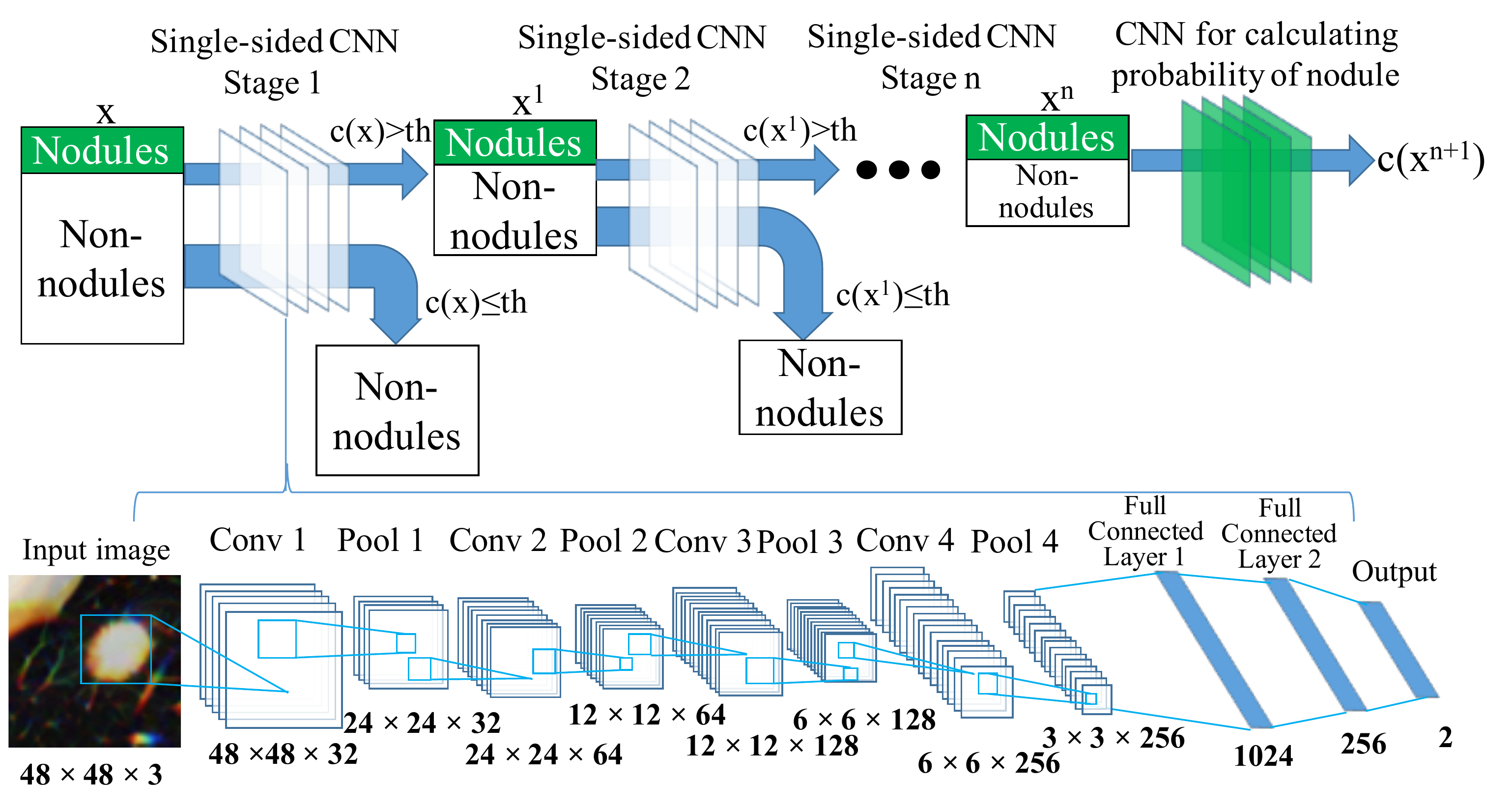}
\caption{Schematic diagram of cascaded multi-stage CNNs. Stage 1, Stage 2 and Stage n are CNNs that perform as single-sided classifiers to filtrer out non-nodule lesions. The final stage is a CNN to calculate nodule probabilities. $c(x)$ is nodule probability of nodule candidate $x$. $th$ is a threshold value to filter out obvious non-nodules. The lower part shows the structure of the CNN. The numbers at lowest part show number of neurons in three dimensions (width, height and channel).}
\label{fig:ccnn}
\end{figure}

Figure \ref{fig:ccnn} shows the schematic diagram of our method. Stage 1, Stage 2 and Stage n are CNNs that perform as single-sided classifiers and gates to filter out low nodule probability samples and pass through the suscicious nodule samples to down stage. The final stage is the CNNs that calculate nodule probabilities. At Stage 1, by using the CNNs that perform as single-sided classifiers, the test dataset is classified, and then, the samples that probabilities fall below a threshold are removed from the test dataset as the obvious non-nodules. The nodule probabilities of removed samples are assigned zero. At Stage 2, the same procedures are applied again to remove further the obvious non-nodules from the test dataset. In the final stage, the CNNs trained by a balanced dataset calculate the probabilities of the remaining suspicious nodule candidates.
The lower part of Figure \ref{fig:ccnn} shows the structure of the CNN. There are three main operations in the CNN: 1) Convolution with rectified linear unit, 2) Pooling or sub-sampling 3) classification by fully connected layer. The input to the CNN is extracted 2-D patches from three consecutive slices of X-ray CT scan images. The convolution layer will compute the output of neurons that are connected to local regions in the input, each computing a dot product between their weights and a small region they are connected to in the input volume. The purpose of convolution is to extract features from the input image. The pooling layer performs a sub-sampling operation along the spatial dimensions (width, height), resulting in size of single channel becomes half of input. The last fully-connected layers will compute the nodule probabilities. The same CNN are used as the single-sided classifiers and the probability calculation at the final stage.

The unique points of our method are that it uses cascaded multi-stage CNNs that perform as single-sided classifiers and uses the inversed imbalanced data as the training data. In contrast, there are some works (e.g. Viola-Jones \cite{viola2001rapid} and Wu et al. \cite{wu2003learning}) using the weak classifiers to construct boosted cascade layer with simple features. Compared with the weak classifiers, convolutional neural network can automatically capture features from the CT images, which can provide higher accuracy for the detection results. As for the cascaded CNN structure, Li et al. \cite{li2015convolutional} have proposed a cascaded CNN structure for face detection. They use 6 CNNs in the cascade including 3 CNNs to detect the face and 3 CNNs to calibrate the bounding box separately. 
The bounding box calibration is not needed in our proposed method. The application of cascade CNN for face detection \cite{qin2016joint, kalinovskii2015compact} and other kind of image feature detection \cite{chen2016mitosis} can also be found in other works. However, class imbalanced problem is not addressed in these works.

\section{Experiments}
\subsection{Lung CT image dataset}
We use the lung CT scan dataset obtained from Lung Nodule Analysis 2016 \cite{Grandchalengeluna16}. This set includes 888 CT scan images along with annotations that were collected during a two-phase annotation process overseen by four experienced radiologists. Each radiologist marked lesions they identified as non-nodule, nodule $<$ 3 mm, and nodule $\geq$ 3 mm. The dataset consists of all nodules $\geq$ 3 mm accepted by at least 3 out of 4 radiologists. The complete dataset is divided into ten subsets to be used for the 10-fold cross-validation. For convenience, the corresponding class label (0 for non-nodule and 1 for nodule) for each candidate is provided. 1,348 lesions are labeled as nodules and the other 551,062 are non-nodule lesions. In this study, center coordinates of each lesion are given. Examples of non-nodule and nodule images in the dataset are given in Figure 3. We use three consecutive slices to obtain volumetric information. Each image size cropped from CT scan images is 48 pixels $\times$ 48 pixels with a central on the nodule candidate.

\begin{figure}[htb]
\centering
\includegraphics[width=\textwidth]{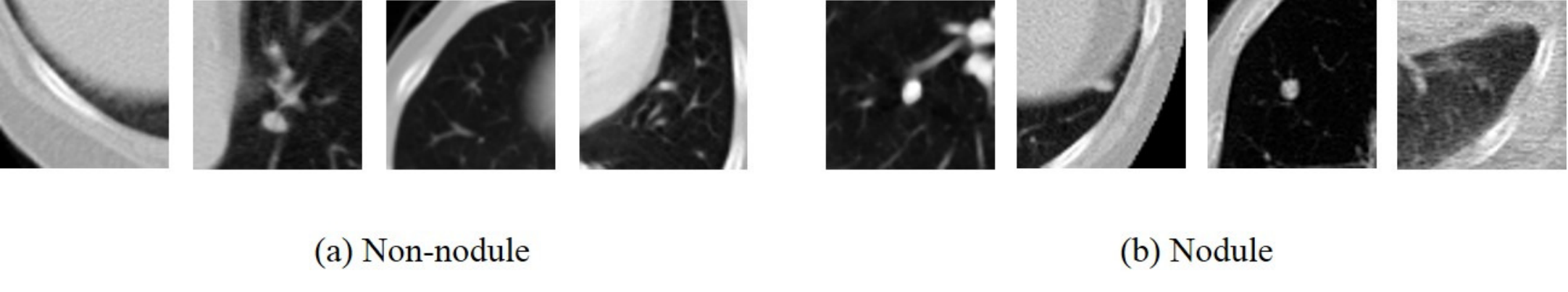}
\caption{Example of lesion images in dataset of Lung Nodule Analysis 2016 \cite{Grandchalengeluna16}.}
\label{fig:samples}
\end{figure}

\subsubsection{Proposed multi-stage classifiers}
The model of single-sided classifiers and the final stage classifiers are trained and validate by 10-fold cross-validation. In the cross-validation, eight subsets are used for training, and one subset is used for calculating accuracy of each models. The remaining subset is used for testing the dataset. 10 CNN models are made by using the holdout procedures. To make the training dataset for the single-sided classifiers, non-nodule samples in a subset are subsampled to 50 samples, and nodule samples are oversampled nine times by randomly rotating and scaling original images. As the result, the number of nodules is about twenty four times the number of non-nodules in the training dataset. In the learning loops of the single-sided classifiers, CNN models having the best nodule classification accuracy over all learning epochs are stored. There are 20 epochs in each training. In the test phase of single-sided classifiers, 
if the probability value of a nodule candidate falls below a specific threshold value, it is classified as an obvious non-nodule, and removed from the subset and assigned zero probability. The threshold value is determined from a standard deviation $\sigma$ of the nodule probability distribution of non-nodule samples. One-tenth of the standard deviation is set as the threshold value. Subsequent stage, the same procedures are repeated for the filtered dataset at the previous stage. At the final stage, the CNN trained by a balanced dataset extracted from the filtered dataset at the previous stage. The CNN models having the best classification accuracy (nodules and non-nodules) over all learning epochs are stored and calculate the probabilities of the nodule candidates of the filtered dataset at the previous stage.

\subsubsection{Baseline classifiers}
For performance comparison, the same CNNs is trained and tested by using the same dataset in manner of the 10-fold cross-validation. We call this conventional method as “baseline”. The CNNs are trained using a balanced dataset with subsampled non-nodules and oversampled nodules. The all nodule samples are oversampled nine times by randomly rotating and scaling original images and non-nodules are subsampled to balance the number of nodule samples. In the training phase, the CNN models having the best classification accuracy over all learning epochs are stored.

\begin{figure}[htb]
\centering
\includegraphics[width=\textwidth]{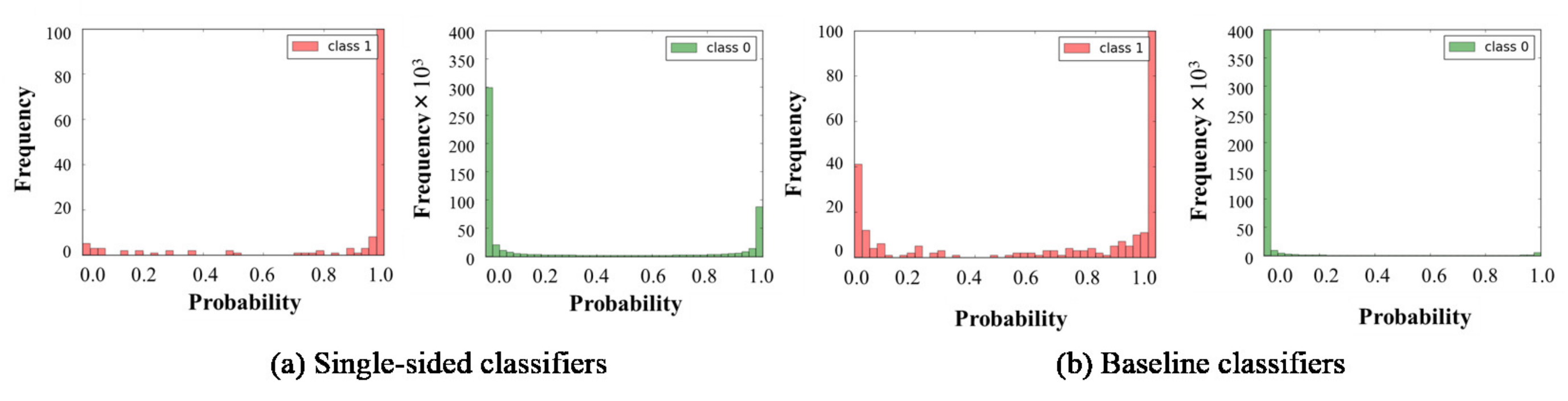}
\caption{Histogram of nodule probabilities of the data set. class 0 is non-nodules class and class 1 is nodule class.}
\label{fig:bl-ss-histos}
\end{figure}

\section{Experimental results}
Figure \ref{fig:bl-ss-histos}(a) shows the histogram of nodule probabilities calculated by the single-sided classifiers at the first stage. The probabilities of non-nodule class (class 0) are separated around 0.0 and 1.0. We assume the samples around probability 0.0 can be accepted as obvious non-nodules. At the same time, 
nodule samples (class 1) around probability 0.0 are accidentally classified as obvious non-nodules. This is what causes the false negatives in our method. Figure \ref{fig:bl-ss-histos}(b) shows the histogram of the nodule probabilities calculated by the baseline classifiers. Although most of the non-nodule samples are concentrated around probability 0.0, a little concentration of the nodule samples also seen around 0.0. This is what causes the false positives in the baseline classifiers. The number of nodule samples around probability 0.0 are more than single sided classifiers. This is what cause low sensitivity in the baseline classifiers.

\begin{figure}[htb]
\centering
\includegraphics[width=\textwidth]{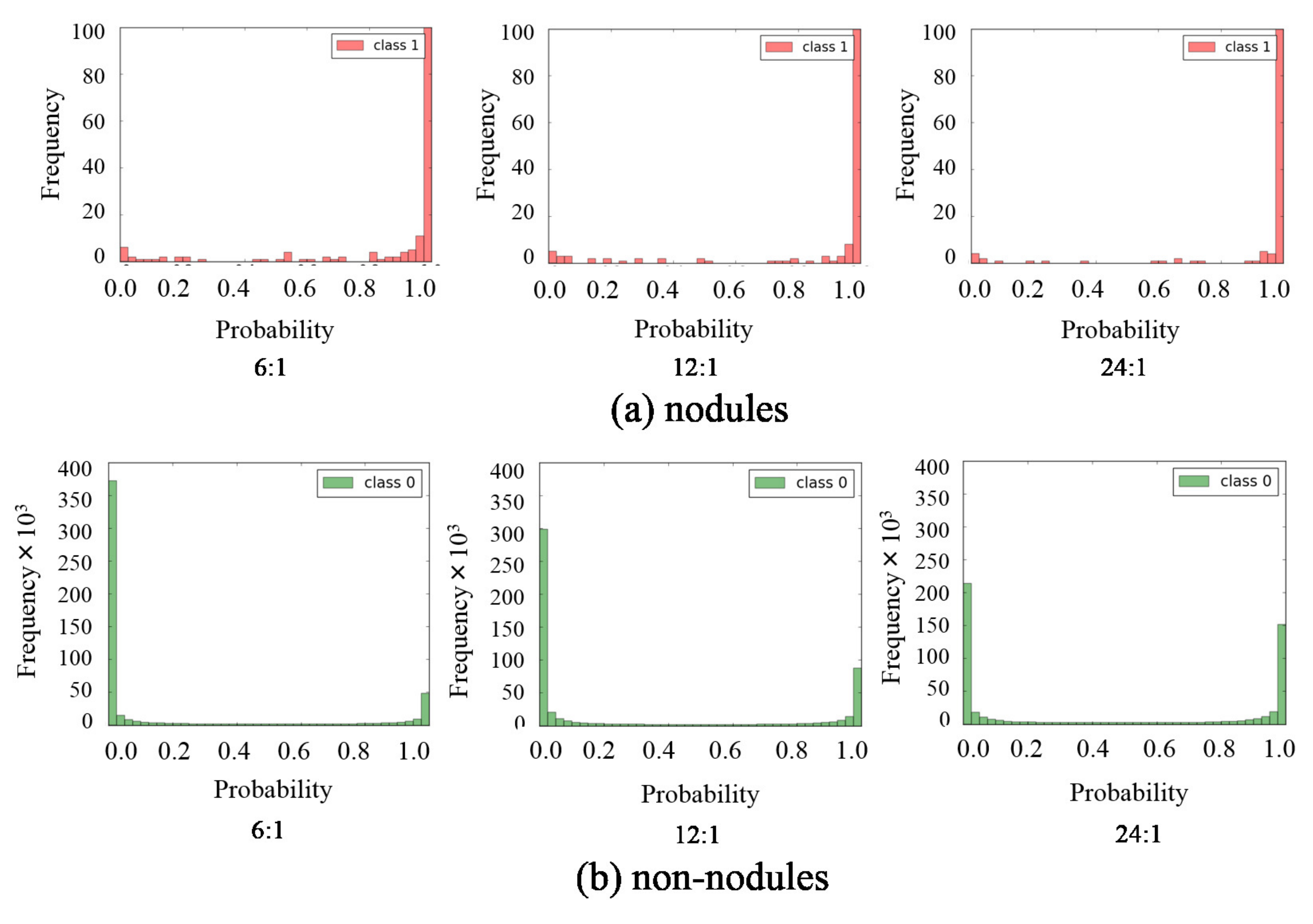}
\caption{Histograms of nodule probabilities of: (a) nodules and (b) non-nodules, calculated by the single-sided classifier at the first stage. Three kind of class sample ratios are compared. The number of nodules to non-nodules are 6 to 1, 12 to 1 and 24 to 1. }
\label{fig:histograms}
\end{figure}

\begin{figure}[htb]
\centering
\includegraphics[width=\textwidth]{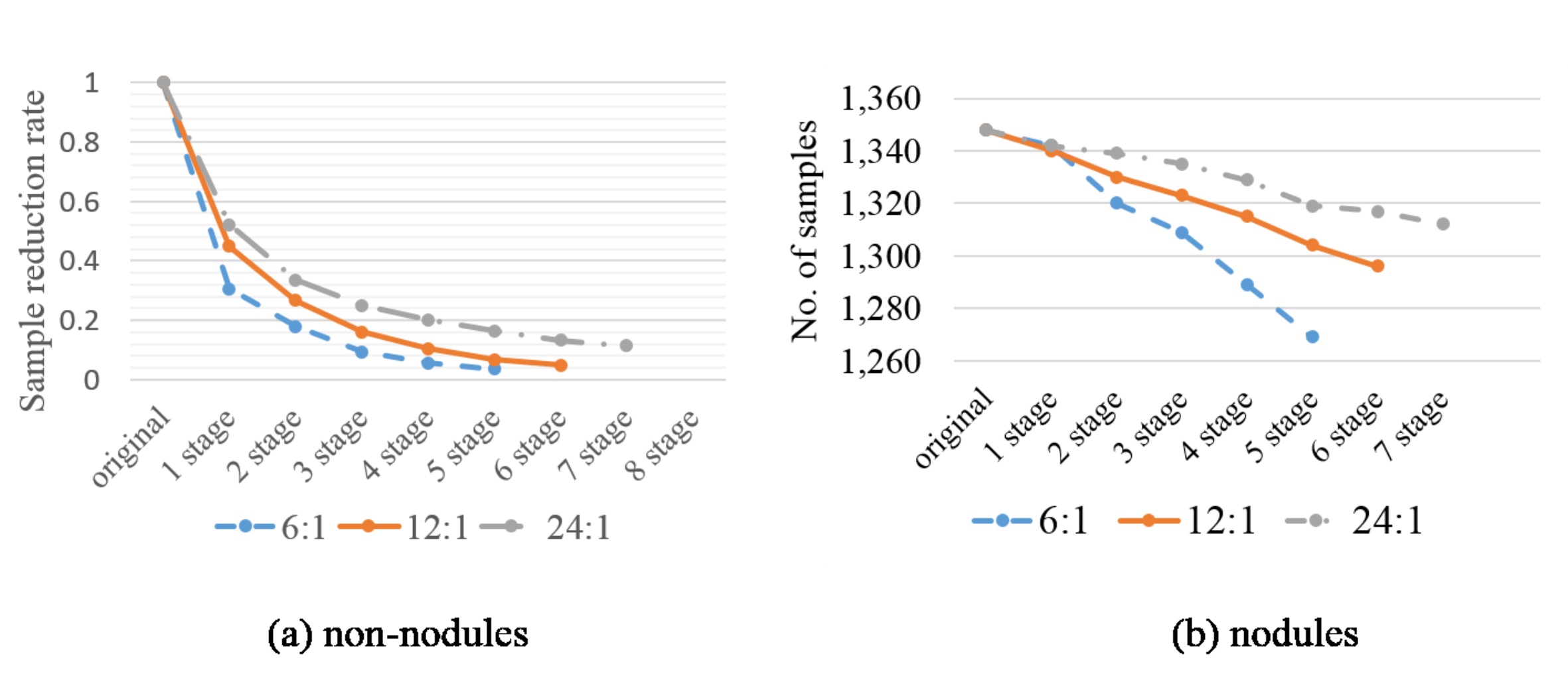}
\caption{Sample reduction rate of non-nodules and decreasing of nodule samples at each stage with deferent class sample ratios.}
\label{fig:screening}
\end{figure}

We investigated the performance of proposed method with the different class sample ratios, the number of nodules to non-nodules are 6 to 1, 12 to 1 and 24 to 1. Figure \ref{fig:histograms} shows histograms of nodule probabilities calculated by the single-sided classifier at the first stage. The nodule samples with small probability decrease as ratio is large as shown in Figure \ref{fig:histograms}(a). At the same time, the non-nodule samples with small probability decrease as ratio is large as shown in Figure \ref{fig:histograms}(b).  As the results, we obtained the non-nodule sample reduction rate at each stage with different class sample ratios as shown in Figure \ref{fig:screening}(a). The number of non-nodules is decrased less than half at the first stage. By cascading the single-sided classifiers and the the obvious non-nodule filtering, the number of non-nodules decreases further. The sample reduction rate of nodule samples reaches under 0.25 at 3 stage. At the same time, the number of nodules are accidentally decreases as stage is later as shown in Figure \ref{fig:screening}(b). This is a side effect of our method. Larger class sample ratio results in slow down the sample reduction. It has a disadvantage for the false positive reduction. However, it is good for the sensitivity because the false negatives decrease. 

\begin{figure}[htb]
\centering
\includegraphics[width=10cm]{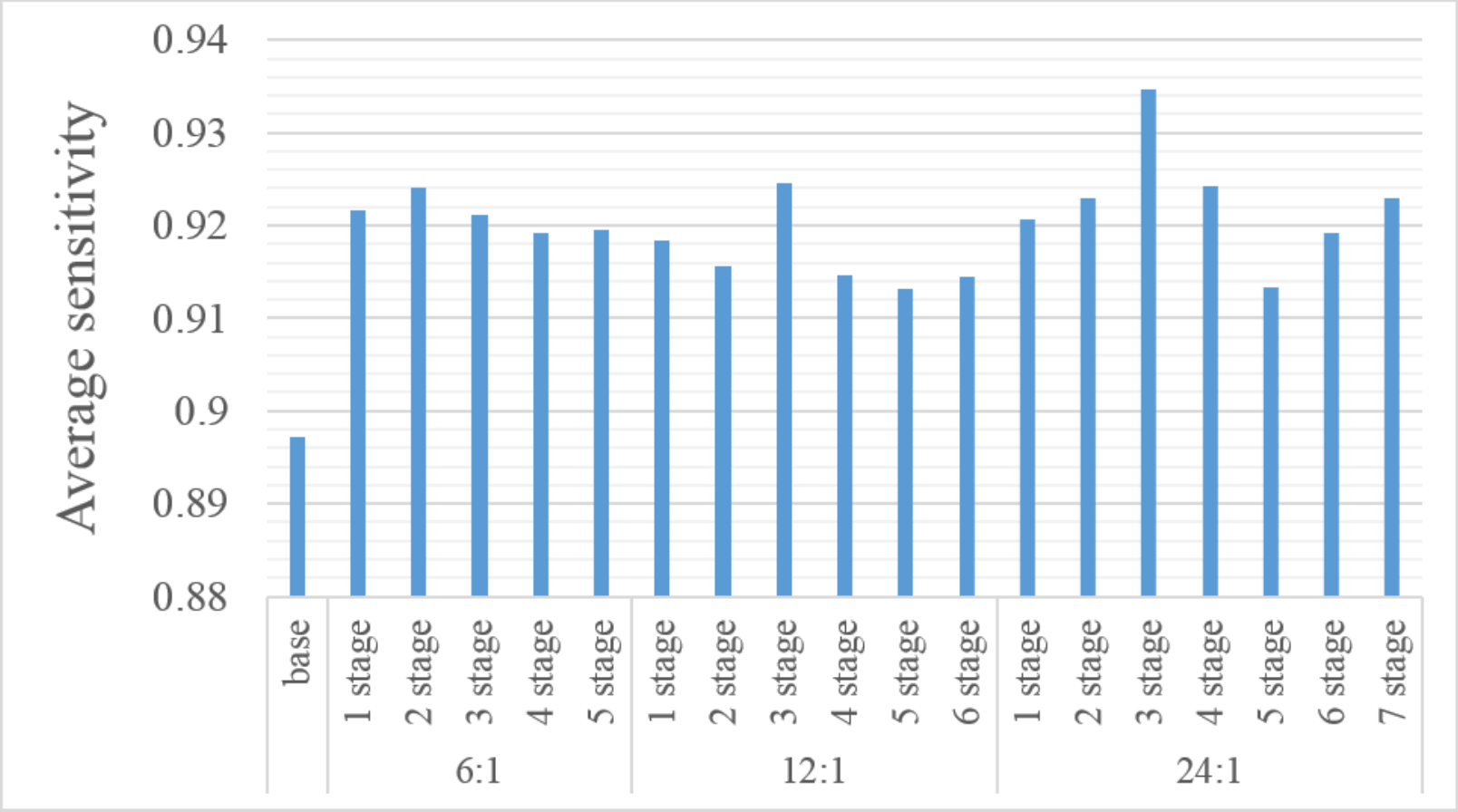}
\caption{Average sensitivity at each stage with different class sample ratio. Three stage single-sided classifiers trained by 24:1 ratio dataset has the best performance.}
\label{fig:ave-perfs}
\end{figure}

\begin{figure}[htb]
\centering
\includegraphics[width=10cm]{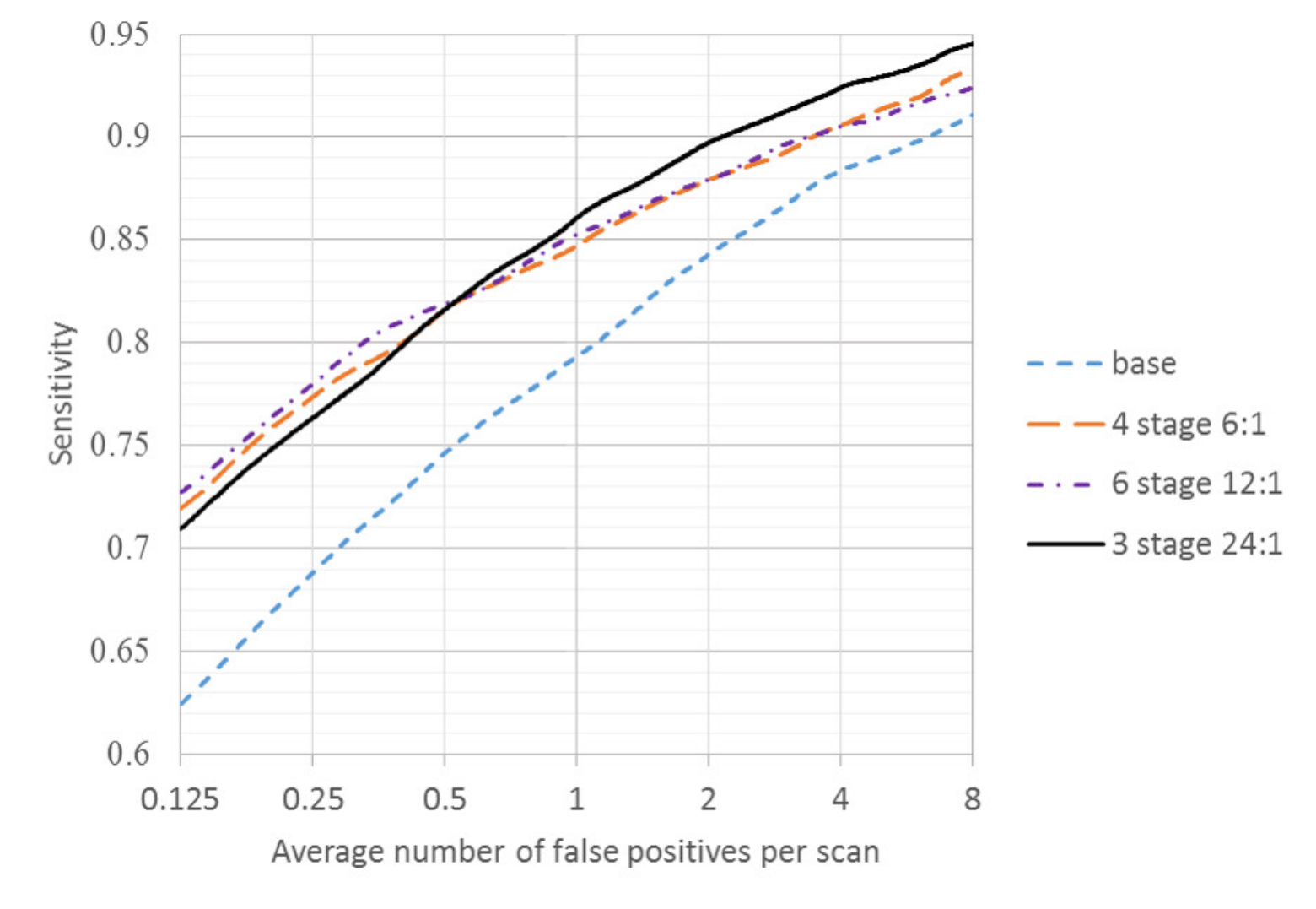}
\caption{FROC curves of proposed methos and baseline.}
\label{fig:froc2}
\end{figure}

Figure \ref{fig:ave-perfs} shows the average sensitivity at each stage with different ratio. The average values are derived from sensitivities at 4 and 8 false positives per scan in FROC curve analysis. In the all cases, our method outperforms the baseline. The best performance is available in 3 stage at 24:1 class sample ratio. Figure \ref{fig:froc2} shows FROC curves at 4 stage with 6:1 ratio, 6 stage with ratio 12:1 and 3 stage with ratio 24:1. The baseline achieves the sensitivity of 88.4\% and 91.1\% at 4 and 8 false positives per scan, respectively. The 3-stage single sided classifiers with 24:1 ratio training reaches the sensitivity of 92.4\% and 94.5\% at 4 and 8 false positives per scan, respectively.

\section{Conclusion}
 In this paper, we have presented cascaded multi-stage neural networks with single-sided classifiers to reduce the false positives of lung nodule classification in CT scan images. We have shown that the proposed method achieves better results in the false positive reduction in comparison with a conventional CNN approach and other approaches. This results present nodule candidates with nodule probabilities to radiologists, which suggests that the system can decrease the burden of image interpretation on radiologists. However, with respect to this detailed theory, there are several unsolved questions. For example, we cannot explain why the average sensitivities of 12:1 and 24:1 fluctuates from stage to stage, and why the 3 stage of 24:1 ratio has the best performance. We will address these questions by the theoretical studies.

\newpage
\bibliographystyle{abbrv}
\bibliography{ICIAP2017-WH-TSDL-IBM}

\end{document}